\documentclass[
]{ceurart}

\usepackage{listings}
\usepackage{booktabs}

\usepackage[whole]{bxcjkjatype} 
\usepackage{ulem}

\begin{document}

\copyrightyear{2025}
\copyrightclause{Copyright for this paper by its authors.
  Use permitted under Creative Commons License Attribution 4.0
  International (CC BY 4.0).}

\conference{This is a preprint version of a paper reviewed and accepted at BREV-RAG 2025: Beyond Relevance-based EValuation of RAG Systems, a SIGIR-AP 2025 workshop.}

\title{RAG System for Supporting Japanese Litigation Procedures: Faithful Response Generation Complying with Legal Norms}

\author[1]{Yuya Ishihara}[
  email=yuya.ishihara@r.hit-u.ac.jp,
  orcid=0009-0003-8033-9927
]
\cormark[1]
\cortext[1]{Corresponding author.}

\author[1]{Atsushi Keyaki}[
  email=a.keyaki@r.hit-u.ac.jp,
  orcid=0000-0001-6495-117X
]
\cormark[1]
\cortext[1]{Corresponding author.}

\author[2]{Hiroaki Yamada}[
    email=yamada@comp.isct.ac.jp,
    orcid=0000-0002-1963-958X
]
\author[1,3]{Ryutaro Ohara}[
    email=r.ohara@ntmlo.com,
    orcid=0009-0000-8018-3895
]
\author[1]{Mihoko Sumida}[
    orcid=0000-0002-8531-2964,
    email=m.sumida@r.hit-u.ac.jp
]

\address[1]{Hitotsubashi University, Japan}
\address[2]{Institute of Science Tokyo, Japan}
\address[3]{Nakamura, Tsunoda \& Matsumoto, Japan}

\begin{abstract}
This study discusses the essential components that a Retrieval-Augmented Generation (RAG)-based LLM system should possess in order to support Japanese medical litigation procedures complying with legal norms.
In litigation, expert commissioners, such as physicians, architects, accountants, and engineers, provide specialized knowledge to help judges clarify points of dispute.
When considering the substitution of these expert roles with a RAG-based LLM system, the constraint of strict adherence to legal norms is imposed.
Specifically, three requirements arise: (1) the retrieval module must retrieve appropriate external knowledge relevant to the disputed issues in accordance with the principle prohibiting the use of private knowledge, (2) the responses generated must originate from the context provided by the RAG and remain faithful to that context, and (3) the retrieval module must reference external knowledge with appropriate timestamps corresponding to the issues at hand.
This paper discusses the design of a RAG-based LLM system that satisfies these requirements.
\end{abstract}

\begin{keywords}
Retrieval--Augmented Generation \sep
Litigation Procedures \sep
Legal Norms \sep
Expert Knowledge \sep
Information Retrieval
\end{keywords}

\maketitle

\section{Introduction}
In recent years, large language models (LLMs) have demonstrated remarkable advancements in their capabilities, leading to a growing movement toward their implementations into professional domains such as medicine and law.
Since they are trained on extensive large text corpora, LLMs acquire a broad collection of knowledge throughout the training process~\cite{kb, emergent}.
However, LLMs do not retain up-to-date information about events that occurred after their training period, and their knowledge of highly specialized or less common domains is not necessarily adequate.
For these reasons, in professional domains such as legal\cite{legal-rag}  and medicine\cite{med-rag} , recent research has increasingly explored the use of Retrieval-Augmented Generation (RAG) approaches, which exploits external knowledge to generate high-quality responses.

\begin{figure}
  \centering
  \includegraphics[width=0.5\linewidth]{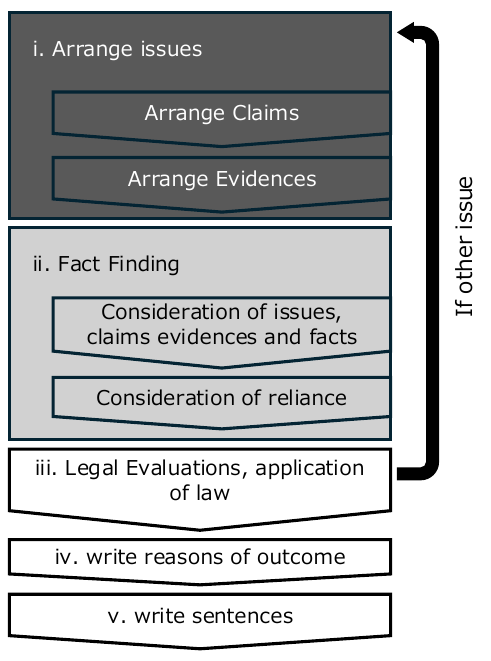}
  \caption{process of civil litigation}
  \label{fig:process}
\end{figure}

Our research group is studying a RAG-based LLM system to support medical litigation procedures in Japan.
Normally, litigation process goes as follows, (i)arranging issues, (ii)fact finding, (iii)legal evaluation, (iv)writing reasons of outcome and (v)writing sentences\ref{fig:process}.
In the process of (i)arrange issues, each claim submitted by the plaintiff and the defendant is examined to distinguish the points of agreement from those in dispute, thereby extracting points that should be the focus points of the litigation.
In medical litigation, this process involves technical advisors who are medical professionals. These experts provide technical explanations regarding matters that may require witness examination or expert testimony during (ii)fact finding, attend sessions involving the discovery of evidence or witnesses, and offer explanations to judges to aid in assessing the reliability of evidence and witness testimonies presented by the parties.
Particularly, because fact-finding by judges in medical litigation requires domain-specific medical expertise, identifying the matters that should be subject to expert testimony demands extensive referencing and analysis of a large volume of legal and medical claims and related documents.
Consequently, the workload in this process is extremely high.
According to Professor Murata Wataru~(Chuo University), a former judge, Japanese courts prepare internal reference materials summarizing the case overview and the issues for expert testimony and opinion to facilitate judicial deliberation.
Thus, support by computers, especially by the application of LLMs and RAG, is expected to significantly enhance the efficiency of these processes.
Furthermore, in a RAG-based LLM system designed to support medical litigation procedures, it is essential not only to provide accurate information but also to ensure compliance with the judicial system's norms.
Based on this premise, we propose a set of requirements that a legal RAG framework should satisfy in order to fulfill those normative principles.

\section{Requirements in medical litigations}
As a preliminary requirement, it is essential to retrieve external knowledge that is appropriate and relevant to the issues of dispute.
Although the accuracy of the retrieval module is one of the established evaluation points in RAG systems, conducting appropriate expert testimony in the context of medical litigation further requires to comply with the norms of civil procedure and consideration of the domain-specific characteristics within the medical litigations.

\subsection{Procedural Requirements for Use of Expert Knowledge}
In Japan's civil litigation procedure, regardless of whether a case involves a specialized domain, the adversarial principle is adopted. Judges make decisions from a neutral point based solely on the claims and evidence submitted by both parties.
Consequently, judges are restricted from relying on issues not submitted or raised by any of the parties or on knowledge not contained within the submitted evidence, as doing so would infringe upon the procedural rights of the parties.
This adversarial principle, has a tension with the expectation of the litigation system that judges continually update their understanding of precedents, statutes, and social norms.
The extent to which judges should be permitted to conduct judicial investigation and use privately acquired knowledge has thus become a subject of debate, particularly in the context of the ongoing digital transformation of society. This requirement applies not only to medical litigations but also to other types of litigation, particularly those that are highly specialized and involve expert advisory systems, such as intellectual property, construction and system development litigations.

Currently, it is recognized that when judges rely on specialized knowledge, domain-specific expert knowledge can be invoked without the presentation of evidence only if it has undergone critical verification by the relevant expert community, and the parties must be guranteed an opportunity to contest the use of such knowledge\cite{civil_litigation1}. Furthermore, in the civil procedure systems of the United States, the United Kingdom, and Germany, the use of privately acquired expert knowledge is considered to be permitted under certain requirements, such as it is being commonly shared within the relevant expert community, or the implementation of procedural requirements including disclosure to both parties and the provision of an opportunity for comment\cite{civil_litigation2}.
Accordingly, in the context of RAG systems, the external knowledge sources should be limited to those that have been critically validated by expert communities, and access to such information must be controlled to ensure that it remains equally available to both parties involved in the proceedings.

\subsection{Reliance of Expert Knowledge and Frequent Updates}
Judges, by their professional responsibilities, are required to continually update their understanding of statutes, judicial precedents, and social norms. In the medical litigation, moreover, judges are additionally expected not only to address the specialized nature of the cases in charge but also to ensure the reliability of the data and evidence upon which their judgments rely.
Furthermore, doctors who serve as technical advisors in medical domains provide technical explanations grounded in their medical expertise, and also update them.
Due to continuous advances in medicine, even authoritative data sources such as well-established medical textbooks, peer-reviewed papers, and clinical guidelines issued by professional communities gradually lose their validity over time.
According to Professor Shigeto Yonemura~(The University of Tokyo), a leading authority in Japanese medical law and also a doctor, approximately twenty percent of such data becomes outdated within five years.
While the knowledge acquired through pretraining inevitably becomes obsolete, continuously retraining LLMs to update domain-specific knowledge would be an inefficient approach.
Therefore, it is essential that generated responses explicitly derive from and be faithful to the context provided through retrievals.

\subsection{Issue-specific Reference Time}
One of the reasons why expert testimony in arranging issues can become a complex procedure is that the applicable standard of expert knowledge differs depending on the issue in dispute.
For example, when the issue concerns whether a physician was negligent or not, the judgment must be based on the medical knowledge, standard of care, and medical law valid at the time the incident happened.
In contrast, when the issue concerns the causal relationship between a medical treatment and its outcome, the judgment should rely on the most up-to-date knowledge available at the close of oral proceedings\cite{ju}.
Therefore, it is necessary to reference external knowledge corresponding to the appropriate time period relevant to the issue in focus.

Although authoritative data gradually lose their validity over time, the transition to new authorized knowledge does not occur abruptly at once.
During the transitional period, multiple streams of expert knowledge that contradict each other may coexist until new data or precedents become widely acknowledged.
Therefore, when retrieving appropriate sources in a RAG framework, it is necessary to consider the expert knowledge that was valid at the relevant point in time.

Based on the above, this study addresses the realization of a ``norm-compliant RAG'' system, focusing on: (1)controlling knowledge sources in compliance with procedural requirements concerning the use of expert knowledge, (2)attribution and faithfulness of generated responses
to their information sources; and (3)appropriateness of the published time of referenced sources.

\section{Related Work}
\subsection{Retrieval-Augmented Generation (RAG)}
\begin{figure}
  \centering
  \includegraphics[width=0.8\linewidth]{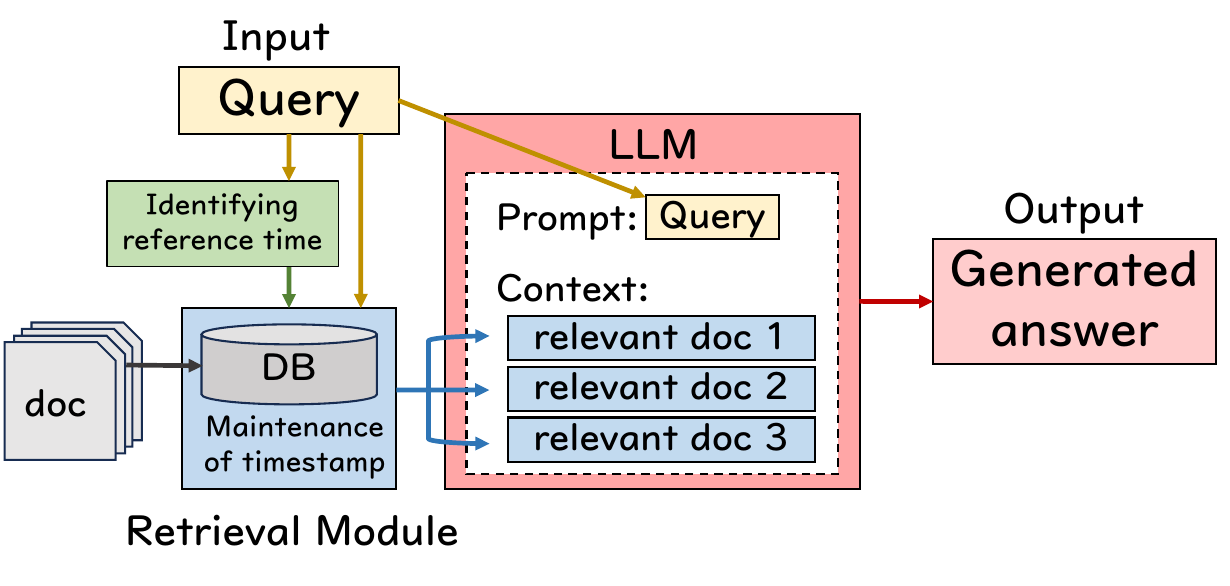}
  \caption{Overview of the a RAG-based LLM System}
  \label{fig:rag}
\end{figure}
Since large language models (LLMs) are trained on extensive corpora, they acquire various forms of knowledge during the learning process~\cite{kb, emergent}.
However, they do not retain up-to-date information, such as current events that occur after model training, and their coverage of specialized or less common knowledge is often insufficient.
As a result, LLMs are known to generate responses containing misinformation, commonly referred to as \textit{hallucinations}.
Because the training corpora used in constructing LLMs may not adequately include domain-specific expertise, the presence of hallucinations is particularly likely when applying such models to specialized domains.

Several approaches have been proposed to mitigate hallucinations, including improving the quality of training data~\cite{data-selection}, adjusting decoding strategies~\cite{inference-time}, enabling self-verification by the model~\cite{selfcheckgpt, self-alignment}, and regenerating responses based on factual verification results~\cite{factcheck-bench}.
Among these, Retrieval-Augmented Generation (RAG)~\cite{rag-survey-1, rag-survey-2} has emerged as one of the most prominent and widely studied approaches.

An overview of the RAG framework is presented in Figure~\ref{fig:rag}.
First, the user's input to the LLM is used as a query to retrieve relevant documents through a retrieval module.
The retrieved documents are then provided to the LLM as contextual information, together with the user's input.
The LLM generates a response while referring to these relevant documents.
By leveraging high-quality external information through RAG, previous studies have reported improvements in task performance~\cite{rag-survey-1, rag-survey-2} and reductions in hallucination occurrence~\cite{hal-mitigate, rag-mit-hal1, rag-mit-hal2, legal-rag, med-rag}.

However, completely suppressing hallucinations remains challenging even when using RAG.
For example, a study on the application of RAG in the legal domain~\cite{legal-rag} reported that, although hallucinations can be mitigated through RAG, they cannot be entirely eliminated.
Consequently, the study emphasizes the importance of expert responsibility in verifying the texts generated by LLMs when applying AI within the legal field.

In addition, \cite{legal-rag} conducted an evaluation of hallucinations based on accuracy and factuality.
Therefore, to the best of our knowledge, no prior research has focused on compliance with legal norms or on the appropriateness of the knowledge sources that substantiate such compliance, which constitutes the central challenge addressed in this study.

\subsection{Analysis of the Correspondence Between Information Sources and Responses}
To verify whether the responses are faithfully generated based on the context provided by the RAG system, possible approaches include analysis using Data Attribution (DA) and evaluation methods related to response faithfulness.

In existing studies on Data Attribution (DA)~\cite{da-1, da-2}, the focus of analysis has been on the pre-training data of LLMs, known as Training Data Attribution (TDA).
In contrast, in the RAG-based LLM system examined in this study, knowledge derived from the RAG component and that originating from the LLM's pre-training data may exist in a competitive relationship, thereby requiring a more complex analytical approach.

Additionally, within the RAG framework, mechanisms have been proposed to evaluate the faithfulness of responses with respect to the provided context.
For example, Ragas\footnote{\url{https://docs.ragas.io/en/stable/}} enables the computation of a Faithfulness Score, which assesses the degree of consistency between the context and the generated response.
The Faithfulness Score determines, through natural language inference, whether the content of the generated response is supported by the information contained in the given context.
Specifically, the Faithfulness Score is calculated through the following procedure:
\begin{enumerate}
  \item Identify all the claims in the response.
  \item Check each claim to see if it can be inferred from the retrieved context.
  \item Compute the faithfulness score using the formula:
\begin{equation}
\text{Faithfulness Score} = 
\frac{\text{Number of claims in the response supported by the retrieved context}}
     {\text{Total number of claims in the response}}
\end{equation}
\end{enumerate}

\section{Norm-compliant RAG}
In this section, we discuss: (1) controlling knowledge sources in compliance with procedural requirements concerning the use of expert knowledge; (2) attribution and faithfulness of generated responses to their information sources; and (3) appropriateness of the published time of referenced sources, to realize a norm-compliant RAG.

\subsection{Controlling Knowledge Sources in Compliance with Procedural Requirements}
We restrict the use of external knowledge to sources that are acceptable according to civil litigation norms.
We can achieve this control by controlling the scope of documents targeted by RAG retrieval and filtering the results.
In assessing this aspect, we could simply label outputs derived from sources that deviate from the predefined scope as inappropriate.

\subsection{Attribution and Faithfulness of Generated Responses to Information Sources}
Expert knowledge is continuously updated over time. Thus, responses generated by relying solely on the model's knowledge acquired during its pre-training period can become easily outdated. RAG is the solution for this issue. To reinforce the effect of RAG, it is necessary to devise methods that generate responses faithful to the context (or documents) retrieved in the pipeline of the RAG approach. Possible approaches include explicit constraints through prompting and the introduction of chain-of-verification steps that check whether candidate sentences for generation are contained in the context.

To assess this aspect, we need to identify the data source or authority via DA analysis. We check whether the information contained in the outputs originates from RAG-derived knowledge or from pre-training. If it originates from pre-training, it is regarded as an inappropriate answer.
Even if the generated output is based on RAG-derived knowledge, if it contradicts the knowledge in the source, it should be considered inappropriate. Thus, it is also necessary to assess the faithfulness of the outputs against the source. 

\subsection{Valid Time Period of Referenced Sources}
The older documents can be outdated if they are overruled due to new discoveries and updates. A naive solution to this issue might be keeping the knowledge always updated to the latest version; however, this solution would not work in our legal RAG setting.

The valid time periods of information sources vary depending on the types of issues raised in trials. For instance, if the issue of interest in a trial is a physician's negligence, the medical knowledge valid at the time of the physician's act can differ from that which is valid at the time of the trial, which is based on newer sources. 
If a system generates responses only according to the newest sources, it makes up an unrealistic conclusion based on knowledge unavailable at the time of the act in question.
Moreover, the validity of time periods for sources matters not only in medical expert knowledge but also in legal expert knowledge, such as precedents and statutes. 

Thus, a legal RAG system should be able to recognize and manage the validity of time periods for sources correctly. Managing the time metadata of information sources is important, especially concerning when information is published and when it becomes invalid. We could achieve this by utilizing timestamps and citation networks.

When assessing this aspect, if a generated response is based on information with inappropriate timestamps that do not align with the input query, it is considered unsuitable.

\section{Conclusion}
We are developing a RAG-based LLM system to support medical litigation proceedings in Japan. Such a system must not just present accurate information, but also provide legally compliant responses to support expert testimony.
To accommodate the requirements, we propose aspects of ``conformance to the norm'' that a legal RAG system should satisfy.  
Specifically, we propose three aspects: (1) controlling knowledge sources in compliance with procedural requirements concerning the use of expert knowledge; (2) attribution and faithfulness of generated responses to their information sources; and (3) appropriateness of the published time of referenced sources.

Our future work includes proposing methods that satisfy each requirement, refining evaluation metrics, implementing them, and conducting experiments.

\section*{Acknowledgments}
We appreciate Prof. Shigeto Yonemura (The University of Tokyo), Prof. Wataru Murata (Chuo University), Prof. Shozo Ota (Meiji University), Prof. Simon Deakin (University of Cambridge), and Prof. Felix Steffek (University of Cambridge) for their helpful comments. This work was partially supported by Minji-Funsou-Shori-Kenkyukikin, the Japanese Society for the Promotion of Science Grantin-Aid for Scientific Research (B) (\#23H03686, \#25K03178) and Scientific Research (C) (\#24K15066), and JST PRESTO (\#JPMJPR236B).

\section*{Declaration on Generative AI}
During the preparation of this work, the authors used GPT-5 and Grammarly for grammar and style suggestions.
After using this tool, the authors reviewed and edited the content and take full responsibility for the publication's content.

\bibliography{reference}

\begin{thebibliography}{19}
\expandafter\ifx\csname natexlab\endcsname\relax\def\natexlab#1{#1}\fi
\providecommand{\url}[1]{\texttt{#1}}
\providecommand{\href}[2]{#2}
\providecommand{\path}[1]{#1}
\providecommand{\DOIprefix}{doi:}
\providecommand{\ArXivprefix}{arXiv:}
\providecommand{\URLprefix}{URL: }
\providecommand{\Pubmedprefix}{pmid:}
\providecommand{\doi}[1]{\href{http://dx.doi.org/#1}{\path{#1}}}
\providecommand{\Pubmed}[1]{\href{pmid:#1}{\path{#1}}}
\providecommand{\bibinfo}[2]{#2}
\ifx\xfnm\relax \def\xfnm[#1]{\unskip,\space#1}\fi
\bibitem[{Petroni et~al.(2019)Petroni, Rockt{\"a}schel, Riedel, Lewis, Bakhtin, Wu, and Miller}]{kb}
\bibinfo{author}{F.~Petroni}, \bibinfo{author}{T.~Rockt{\"a}schel}, \bibinfo{author}{S.~Riedel}, \bibinfo{author}{P.~Lewis}, \bibinfo{author}{A.~Bakhtin}, \bibinfo{author}{Y.~Wu}, \bibinfo{author}{A.~Miller},
\newblock \bibinfo{title}{{Language Models as Knowledge Bases?}},
\newblock in: \bibinfo{booktitle}{Proc. of the EMNLP-IJCNLP 2019}, \bibinfo{year}{2019}.
\bibitem[{Wei et~al.(2022)Wei, Tay, Bommasani, Raffel, Zoph, Borgeaud, Yogatama, Bosma, Zhou, Metzler, Chi, Hashimoto, Vinyals, Liang, Dean, and Fedus}]{emergent}
\bibinfo{author}{J.~Wei}, \bibinfo{author}{Y.~Tay}, \bibinfo{author}{R.~Bommasani}, \bibinfo{author}{C.~Raffel}, \bibinfo{author}{B.~Zoph}, \bibinfo{author}{S.~Borgeaud}, \bibinfo{author}{D.~Yogatama}, \bibinfo{author}{M.~Bosma}, \bibinfo{author}{D.~Zhou}, \bibinfo{author}{D.~Metzler}, \bibinfo{author}{E.~H. Chi}, \bibinfo{author}{T.~Hashimoto}, \bibinfo{author}{O.~Vinyals}, \bibinfo{author}{P.~Liang}, \bibinfo{author}{J.~Dean}, \bibinfo{author}{W.~Fedus},
\newblock \bibinfo{title}{Emergent abilities of large language models},
\newblock \bibinfo{journal}{Transactions on Machine Learning Research (TMLR)}  (\bibinfo{year}{2022}) \bibinfo{pages}{2835--8856}.
\bibitem[{Magesh et~al.(2025)Magesh, Surani, Dahl, Suzgun, Manning, and Ho}]{legal-rag}
\bibinfo{author}{V.~Magesh}, \bibinfo{author}{F.~Surani}, \bibinfo{author}{M.~Dahl}, \bibinfo{author}{M.~Suzgun}, \bibinfo{author}{C.~D. Manning}, \bibinfo{author}{D.~E. Ho},
\newblock \bibinfo{title}{{Hallucination‐Free? Assessing the Reliability of Leading AI Legal Research Tools}},
\newblock \bibinfo{journal}{Journal of Empirical Legal Studies} \bibinfo{volume}{22} (\bibinfo{year}{2025}) \bibinfo{pages}{216--242}.
\bibitem[{Chu et~al.(2025)Chu, Zhang, Malon, and Min}]{med-rag}
\bibinfo{author}{Y.-W. Chu}, \bibinfo{author}{K.~Zhang}, \bibinfo{author}{C.~Malon}, \bibinfo{author}{M.~R. Min},
\newblock \bibinfo{title}{{Reducing Hallucinations of Medical Multimodal Large Language Models with Visual Retrieval-Augmented Generation }},
\newblock in: \bibinfo{booktitle}{Proc. of the AAAI 2025}, \bibinfo{year}{2025}.
\bibitem[{Okanari(2021)}]{civil_litigation1}
\bibinfo{author}{G.~Okanari},
\newblock \bibinfo{title}{Saibankan no shichi riyou no kinshi (the prohibition of judge’s use of private knowledge)},
\newblock \bibinfo{journal}{Hougaku Zasshi :(Journal of Law) of Osaka City University} \bibinfo{volume}{68} (\bibinfo{year}{2021}) \bibinfo{pages}{1 -- 66}.
\bibitem[{Sugiyama(2023)}]{civil_litigation2}
\bibinfo{author}{E.~Sugiyama},
\newblock \bibinfo{title}{Saibankan niyoru senmonchishiki no shushu to riyou (collection and use of expert knowledge by the judge, symposium: The discipline of civil judges in the exercise of their powers)},
\newblock \bibinfo{journal}{Minso Zasshi (Journal of Civil Procedure)} \bibinfo{volume}{69} (\bibinfo{year}{2023}) \bibinfo{pages}{103 -- 114}.
\bibitem[{Shirai(2006)}]{ju}
\bibinfo{author}{Y.~Shirai},
\newblock \bibinfo{title}{{Mijukuji Moumakusyou to Ishi no Kashitu (Misdiagnosis of retinopathy of prematurity and Doctor’s Negligence)}},
\newblock \bibinfo{journal}{Hanrei kara Manabu Minji-Jijitsu Nintei: Jurisuto Zoukan (Special Edition of Journal: Jurist: Learning from Case Law: Civil Fact-Finding)}  (\bibinfo{year}{2006}) \bibinfo{pages}{252--256}.
\bibitem[{Albalak et~al.(2024)Albalak, Elazar, Xie, Longpre, Lambert, Wang, Muennighoff, Hou, Pan, Jeong, Raffel, Chang, Hashimoto, and Wang}]{data-selection}
\bibinfo{author}{A.~Albalak}, \bibinfo{author}{Y.~Elazar}, \bibinfo{author}{S.~M. Xie}, \bibinfo{author}{S.~Longpre}, \bibinfo{author}{N.~Lambert}, \bibinfo{author}{X.~Wang}, \bibinfo{author}{N.~Muennighoff}, \bibinfo{author}{B.~Hou}, \bibinfo{author}{L.~Pan}, \bibinfo{author}{H.~Jeong}, \bibinfo{author}{C.~Raffel}, \bibinfo{author}{S.~Chang}, \bibinfo{author}{T.~Hashimoto}, \bibinfo{author}{W.~Y. Wang}, \bibinfo{title}{{A Survey on Data Selection for Language Models}}, \bibinfo{howpublished}{arXiv:2402.16827}, \bibinfo{year}{2024}.
\bibitem[{Li et~al.(2023)Li, Patel, Vi^^c3^^a9gas, Pfister, and Wattenberg}]{inference-time}
\bibinfo{author}{K.~Li}, \bibinfo{author}{O.~Patel}, \bibinfo{author}{F.~Vi^^c3^^a9gas}, \bibinfo{author}{H.~Pfister}, \bibinfo{author}{M.~Wattenberg},
\newblock \bibinfo{title}{{Inference-Time Intervention: Eliciting Truthful Answers from a Language Model}},
\newblock in: \bibinfo{booktitle}{Proc. of the NeurIPS 2023}, \bibinfo{year}{2023}.
\bibitem[{Manakul et~al.(2023)Manakul, Liusie, and Gales}]{selfcheckgpt}
\bibinfo{author}{P.~Manakul}, \bibinfo{author}{A.~Liusie}, \bibinfo{author}{M.~Gales},
\newblock \bibinfo{title}{{SelfCheckGPT: Zero-Resource Black-Box Hallucination Detection for Generative Large Language Models}},
\newblock in: \bibinfo{booktitle}{Proc. of the EMNLP 2023}, \bibinfo{year}{2023}.
\bibitem[{Zhang et~al.(2024)Zhang, Peng, Tian, Zhou, Jin, Song, Mi, and Meng}]{self-alignment}
\bibinfo{author}{X.~Zhang}, \bibinfo{author}{B.~Peng}, \bibinfo{author}{Y.~Tian}, \bibinfo{author}{J.~Zhou}, \bibinfo{author}{L.~Jin}, \bibinfo{author}{L.~Song}, \bibinfo{author}{H.~Mi}, \bibinfo{author}{H.~Meng},
\newblock \bibinfo{title}{{Self-Alignment for Factuality: Mitigating Hallucinations in LLMs via Self-Evaluation}},
\newblock in: \bibinfo{booktitle}{Proc. of the ACL 2024}, \bibinfo{year}{2024}.
\bibitem[{Wang et~al.(2024)Wang, Reddy, Mujahid, Arora, Rubashevskii, Geng, Afzal, Pan, Borenstein, Pillai, Augenstein, Gurevych, and Nakov}]{factcheck-bench}
\bibinfo{author}{Y.~Wang}, \bibinfo{author}{R.~G. Reddy}, \bibinfo{author}{Z.~M. Mujahid}, \bibinfo{author}{A.~Arora}, \bibinfo{author}{A.~Rubashevskii}, \bibinfo{author}{J.~Geng}, \bibinfo{author}{O.~M. Afzal}, \bibinfo{author}{L.~Pan}, \bibinfo{author}{N.~Borenstein}, \bibinfo{author}{A.~Pillai}, \bibinfo{author}{I.~Augenstein}, \bibinfo{author}{I.~Gurevych}, \bibinfo{author}{P.~Nakov},
\newblock \bibinfo{title}{{Factcheck-Bench: Fine-Grained Evaluation Benchmark for Automatic Fact-checkers}},
\newblock in: \bibinfo{booktitle}{Proc. of the Findings of the EMNLP 2024}, \bibinfo{year}{2024}.
\bibitem[{Lewis et~al.(2020)Lewis, Perez, Piktus, Petroni, Karpukhin, Goyal, K^^c3^^bcttler, Lewis, tau Yih, Rockt^^c3^^a4schel, Riedel, and Kiela}]{rag-survey-1}
\bibinfo{author}{P.~Lewis}, \bibinfo{author}{E.~Perez}, \bibinfo{author}{A.~Piktus}, \bibinfo{author}{F.~Petroni}, \bibinfo{author}{V.~Karpukhin}, \bibinfo{author}{N.~Goyal}, \bibinfo{author}{H.~K^^c3^^bcttler}, \bibinfo{author}{M.~Lewis}, \bibinfo{author}{W.~tau Yih}, \bibinfo{author}{T.~Rockt^^c3^^a4schel}, \bibinfo{author}{S.~Riedel}, \bibinfo{author}{D.~Kiela},
\newblock \bibinfo{title}{{Retrieval-Augmented Generation for Knowledge-Intensive NLP Tasks}},
\newblock in: \bibinfo{booktitle}{Proc. of the NIPS 2020}, \bibinfo{year}{2020}.
\bibitem[{Gao et~al.(2023)Gao, Xiong, Gao, Jia, Pan, Bi, Dai, Sun, Wang, and Wang}]{rag-survey-2}
\bibinfo{author}{Y.~Gao}, \bibinfo{author}{Y.~Xiong}, \bibinfo{author}{X.~Gao}, \bibinfo{author}{K.~Jia}, \bibinfo{author}{J.~Pan}, \bibinfo{author}{Y.~Bi}, \bibinfo{author}{Y.~Dai}, \bibinfo{author}{J.~Sun}, \bibinfo{author}{M.~Wang}, \bibinfo{author}{H.~Wang}, \bibinfo{title}{{Retrieval-Augmented Generation for Large Language Models: A Survey}}, \bibinfo{howpublished}{arXiv:2312.10997}, \bibinfo{year}{2023}.
\bibitem[{Tonmoy et~al.(2024)Tonmoy, Zaman, Jain, Rani, Rawte, Chadha, and Das}]{hal-mitigate}
\bibinfo{author}{S.~T.~I. Tonmoy}, \bibinfo{author}{S.~M.~M. Zaman}, \bibinfo{author}{V.~Jain}, \bibinfo{author}{A.~Rani}, \bibinfo{author}{V.~Rawte}, \bibinfo{author}{A.~Chadha}, \bibinfo{author}{A.~Das}, \bibinfo{title}{{A Comprehensive Survey of Hallucination Mitigation Techniques in Large Language Models}}, \bibinfo{howpublished}{arXiv:2401.01313}, \bibinfo{year}{2024}.
\bibitem[{Lewis et~al.(2024)Lewis, Perez, Piktus, Petroni, Karpukhin, Goyal, K^^c3^^bcttler, Lewis, tau Yih, Rockt^^c3^^a4schel, Riedel, and Kiela}]{rag-mit-hal1}
\bibinfo{author}{P.~Lewis}, \bibinfo{author}{E.~Perez}, \bibinfo{author}{A.~Piktus}, \bibinfo{author}{F.~Petroni}, \bibinfo{author}{V.~Karpukhin}, \bibinfo{author}{N.~Goyal}, \bibinfo{author}{H.~K^^c3^^bcttler}, \bibinfo{author}{M.~Lewis}, \bibinfo{author}{W.~tau Yih}, \bibinfo{author}{T.~Rockt^^c3^^a4schel}, \bibinfo{author}{S.~Riedel}, \bibinfo{author}{D.~Kiela},
\newblock \bibinfo{title}{{Reducing Hallucination in Structured Outputs via Retrieval-Augmented Generation}},
\newblock in: \bibinfo{booktitle}{Proc. of the NAACL 2024}, \bibinfo{year}{2024}.
\bibitem[{Zhang and Zhang(2025)}]{rag-mit-hal2}
\bibinfo{author}{W.~Zhang}, \bibinfo{author}{J.~Zhang},
\newblock \bibinfo{title}{{Hallucination Mitigation for Retrieval-Augmented Large Language Models: A Review}},
\newblock \bibinfo{journal}{Mathematics} \bibinfo{volume}{13} (\bibinfo{year}{2025}).
\bibitem[{Pruthi et~al.(2020)Pruthi, Liu, Kale, and Sundararajan}]{da-1}
\bibinfo{author}{G.~Pruthi}, \bibinfo{author}{F.~Liu}, \bibinfo{author}{S.~Kale}, \bibinfo{author}{M.~Sundararajan},
\newblock \bibinfo{title}{{Estimating Training Data Influence by Tracing Gradient Descent}},
\newblock in: \bibinfo{booktitle}{Proc. of the NeurIPS 2020}, \bibinfo{year}{2020}.
\bibitem[{Chang et~al.(2025)Chang, Rajagopal, Bolukbasi, Dixon, and Tenney}]{da-2}
\bibinfo{author}{T.~A. Chang}, \bibinfo{author}{D.~Rajagopal}, \bibinfo{author}{T.~Bolukbasi}, \bibinfo{author}{L.~Dixon}, \bibinfo{author}{I.~Tenney},
\newblock \bibinfo{title}{{Scalable Influence and Fact Tracing for Large Language Model Pretraining}},
\newblock in: \bibinfo{booktitle}{Proc. of the ICLR 2025}, \bibinfo{year}{2025}.

\end{thebibliography}

\end{document}